\documentclass{article}

\usepackage{microtype}
\usepackage{graphicx}
\usepackage{subfigure}
\usepackage{multirow}
\usepackage{colortbl} 
\usepackage{xcolor} 
\usepackage{booktabs} 
\usepackage{makecell}
\usepackage{hyperref}

\usepackage[accepted]{icml2025}

\usepackage{amsmath}
\usepackage{amssymb}
\usepackage{mathtools}
\usepackage{amsthm}
\usepackage{enumitem}
\usepackage{soul}
\usepackage{bbm}
\usepackage[capitalize,noabbrev]{cleveref}

\hypersetup{
    colorlinks,
    linkcolor={red!50!black},
    citecolor={brown!85!red},
    urlcolor={blue!80!black}
}

\theoremstyle{plain}

\theoremstyle{definition}

\theoremstyle{remark}

\newcommand{\adabracket}[1]{\left(#1\right)}
\newcommand{\adarectbracket}[1]{\left[#1\right]}
\newcommand{\expectation}[2]{\mathbb{E}_{#1}\adarectbracket{#2}}

\newcommand{\bz}{\boldsymbol{z}}
\newcommand{\bx}{\boldsymbol{x}}
\newcommand{\ba}{\boldsymbol{a}}
\newcommand{\bs}{\boldsymbol{s}}
\newcommand{\by}{\boldsymbol{y}}
\newcommand{\bq}{\boldsymbol{q}}
\newcommand{\networkMap}[1]{f_{\theta}(#1)}

\definecolor{intnull}{RGB}{213,229,255}
\definecolor{antiquewhite}{rgb}{0.98, 0.92, 0.84}
\definecolor{lightgray}{rgb}{0.83, 0.83, 0.83}
\definecolor{platinum}{rgb}{0.9, 0.89, 0.89}
\usepackage[textsize=tiny]{todonotes}

\newcommand\soutpars[1]{\let\helpcmd\sout\parhelp#1\par\relax\relax}
\icmltitlerunning{Rule-based Reinforcement Learning Layer}

\begin{document}

\twocolumn[
\icmltitle{Towards Physiologically Sensible Predictions \\via the Rule-based Reinforcement Learning Layer}



\icmlsetsymbol{equal}{*}

\begin{icmlauthorlist}
\icmlauthor{Lingwei Zhu}{equal,utokyo}
\icmlauthor{Zheng Chen}{equal,osaka}
\icmlauthor{Yukie Nagai}{utokyo}
\icmlauthor{Jimeng Sun}{uiuc}
\end{icmlauthorlist}

\icmlaffiliation{utokyo}{University of Tokyo, Japan}
\icmlaffiliation{osaka}{Osaka University, Japan}
\icmlaffiliation{uiuc}{University of Illinois at Urbana-Champaign, USA}

\icmlcorrespondingauthor{Lingwei Zhu}{lingwei4@ualberta.ca}
\icmlcorrespondingauthor{Zheng Chen}{chen.zheng.bn1@gmail.com}

\icmlkeywords{Machine Learning, ICML}

\vskip 0.3in
]



\printAffiliationsAndNotice{\icmlEqualContribution} 

\begin{abstract}
This paper adds to the growing literature of reinforcement learning (RL) for healthcare by proposing a novel paradigm: augmenting any predictor with Rule-based RL Layer (RRLL) that corrects the model's physiologically impossible predictions.
Specifically, RRLL takes as input states predicted labels and outputs corrected labels as actions.
The reward of the state-action pair is evaluated by a set of general rules.
RRLL is efficient, general and lightweight: it does not require heavy expert knowledge like prior work but only a set of impossible transitions.
This set is much smaller than all possible transitions; yet it can effectively reduce physiologically impossible mistakes made by the state-of-the-art predictor models.
We verify the utility of RRLL on a variety of important healthcare classification problems and observe significant improvements using the same setup, with only the domain-specific set of impossibility changed.
In-depth analysis shows that RRLL indeed improves accuracy by effectively reducing the presence of physiologically impossible predictions.

\end{abstract}

\section{Introduction}


{\color{black}
The healthcare community has witnessed a surging interest in reinforcement learning (RL) as a solution to a wide variety of  clinical decision problems \citep{Gottesman2019-NatureMed-guidelinesRLinHeathcare,RLCDR,Yu2021-RLHealthcareSurvey}.
Unlike supervised learning models that take as input abundance of IID data and outputs a mapping from data to true labels, 
RL operates under the sequential decision making framework that learns from trial-and-error interaction with the environment \citep{Sutton-RL2018}.
As noted by \citet{Yu2021-RLHealthcareSurvey}, RL is desirable as it is capable of finding optimal policies using only previous experiences of treatment and outcomes. This makes RL more attractive than existing pharmacodynamics- or simulation-based studies where an accurate model is intractable due to heterogeneity of biological systems and their individual responses.
So far, successful RL examples include dynamic adaptive treatment \citep{Guez2008-RLEpilepsyTreatment,Pineau2009-RLEpilepsyStimulationTreatment,Vincent2014-phd-SeizureRL,Fatemi2021-RLidentifyMedicalDeadends} to automatic medical diagnosis \citep{Kao2018-contextAwareDiseaseDiagnosis-RL,Peng2018-FastDiseaseDiagnosisbyRL-Refuel,Xu2019-relationalDialogueAutomaticDiagnosis,Xia2020-GAILPGAutomaticDiagnosis}.
These applications often take as input clinical observations and assessments of patients, with success heavily dwelling on expert knowledge such as clinical decision rules  \cite{CDR2023,Yu2023-DRL-costEffectiveDiagnosis} or dynamics of disease progression \citep{Saboo2021-RLDiseaseProgresionModelAlzheimer,Hu2023-REMEDI-ML4H-RLadaptiveMEtabolismmodeling}.
The expert knowledge is required to design an adequate Markov Decision Process \citep{Puterman1994} such that the states/rewards/dynamics are faithful and informative; the actions are reflective of intended changes and thus the optimal policy can represent meaningful solutions.


\begin{figure}
    \centering
    \includegraphics[width=\columnwidth]{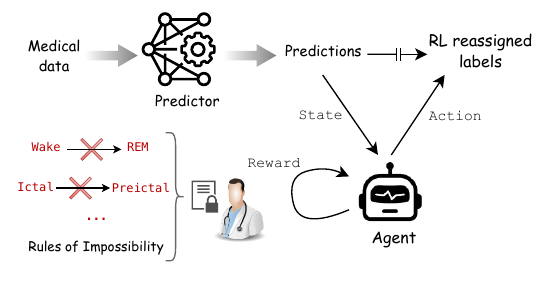}
    \vspace{-15pt}
    \caption{
    Overview of the proposed rule-based reinforcement learning layer. It takes as input state predictions from the predictor and outputs physiological sensible reassigned labels as actions.
    Reward function is jointly defined by the mismatch to true labels and the set of impossible transitions.
    }
    \vspace{-5pt}
    \label{fig:overview}
\end{figure}

This paper complements the growing body of RL for healthcare literature by focusing on another important but less studied problem: supporting physiologically sensible predictions.
This problem naturally emerges from data-driven learning models that achieve high prediction accuracy but can make mistakes that violate the physiological rules.
Take sleep for example, a human goes through five sleep stages, from wake to rapid eye movement (REM).
While a sudden transition from REM to wake is possible, the reverse direction violates the physiological rules and is therefore impossible \cite{AASM}.
However, these rules are implicit underlying training data, and the established techniques or data transformation can cause a discontinuity to the rules.
As a result, the data-driven models on one hand can achieve high accuracy;
on the other hand, they inevitably suffer from physiologically impossible predictions resulting from not respecting the rules during training, preventing them from deployment to real circumstances \cite{PostProcess2,PostProcess1,AAAIConSequence}.

We propose to tackle this challenge with a new framework that augments the data-driven models with an additional Rule-based RL Layer (RRLL).
Specifically, RRLL is simple and lightweight: it does not require heavy expert knowledge such as the knowledge of disease progression to characterize possible states,  dynamics and reward functions \citep{Fatemi2021-RLidentifyMedicalDeadends,Hu2023-REMEDI-ML4H-RLadaptiveMEtabolismmodeling}.
By contrast, it is sufficient for RRLL to function knowing only the rules of impossibility.
This often creates a much smaller set compared to the rules of possibility, e.g. the wake stage can transition to many other stages that eventually lead to REM.
With the rules of impossibility, the RL layer takes in predictions output by a classifier as states, and outputs as actions physiologically sensible, reassigned labels, evaluated by the rules as rewards.
After training, the entire model can work in an end-to-end manner with RRLL simply acting as the last layer.
}

The contributions of the paper are threefold.
Firstly, we present a new framework that augments any data-driven model with an additional rule-based RL layer (RRLL) to support clinically sensible decisions. 
We are unaware of any published results that address a similar idea.
In Section \ref{sec:related_work} we discuss physiological considerations and existing methods in detail.
Secondly, we show the efficacy of the proposed RL layer by extensive experiments with real physiological data for sleep stage prediction and seizure detection.
We comprehensively discuss the MDP for the RL problem in Section \ref{sec:proposed}.
Lastly, we make public the datasets and to facilitate future research such as utilizing more sophisticated RL algorithms for the layer or developing other healthcare datasets into RL environments.

\begin{figure*}[t]
    \centering
    \includegraphics[width=0.96\textwidth]{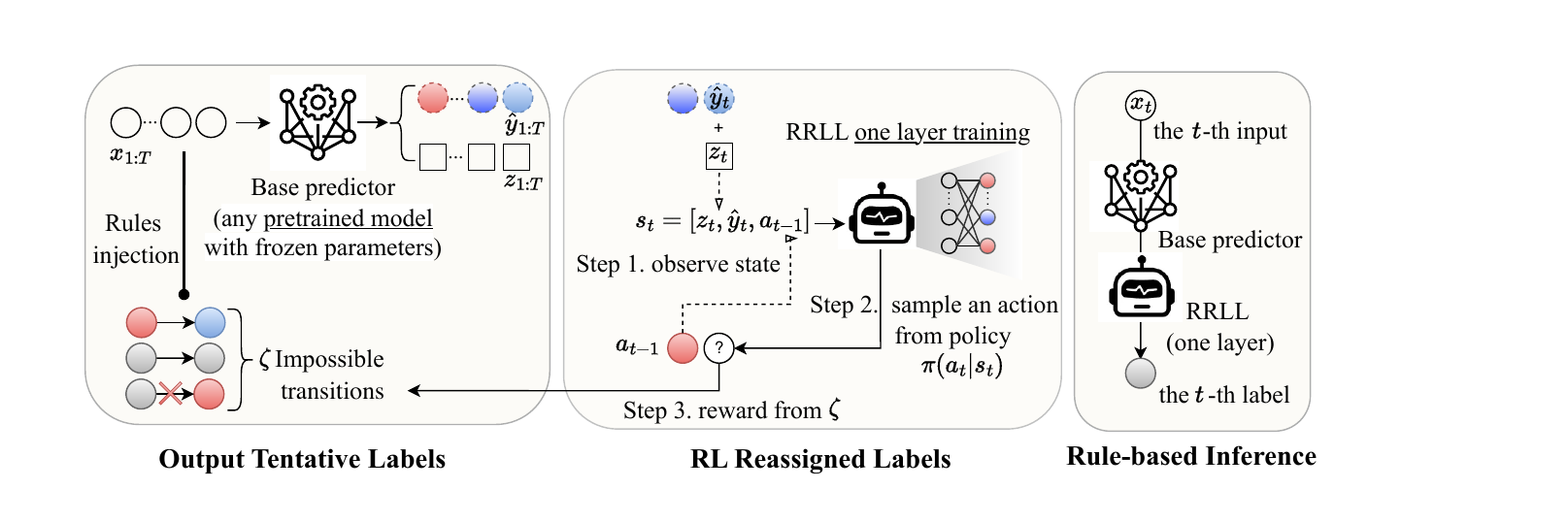}
    \caption{
    Overview of RRLL workflow. 
    (Left) Given a trained base predictor with frozen weights, we loop through the dataset to obtain its tentative predicted labels $\hat{y}$ and features $\bz$. The predicted labels and feature vectors, together with the agent's last action constitute states.
    (Middle) The agent draws an action which is a reassigned label to the input instance. The state-action pair is then evaluated by the set of impossibility $\zeta$.
    (Right) After RRLL learning, the entire model can be used in an end-to-end manner.
    }
    \label{fig:workflow}
\end{figure*}

\section{Related Works}\label{sec:related_work}

\textbf{RL for Healthcare. }
The literature of RL for healthcare has mainly evolved around  dynamic adaptive treatment and automatic diagnosis \citep{Yu2021-RLHealthcareSurvey}. The former being more intuitive, often leverage disease assessments as states and dosage as actions, and the goal is to learn an optimal policy that can adaptively recommend dosage along disease progression \citep{Guez2008-RLEpilepsyTreatment,Pineau2009-RLEpilepsyStimulationTreatment,Fatemi2021-RLidentifyMedicalDeadends,Li2023-quasiOptimal-qGaussian}.
The latter typically abstracts symptoms as a binary state vector, with actions corresponding to queries for a certain symptoms in the state vector. The optimal policy is one that can identify the user's disease as soon as possible by making as few as possible queries \citep{Peng2018-FastDiseaseDiagnosisbyRL-Refuel,Xu2019-relationalDialogueAutomaticDiagnosis,Zhong-2022-RLAutomaticDiagnosis,Yu2023-DRL-costEffectiveDiagnosis}.
Our work differs from them in that RRLL aims at supervising the base predictor model rather than recommending treatments or potential disease classification. 
RRLL can be seen as playing the role of the predictor by drawing reassigned labels, and hence it is really an additional \emph{layer} to the predictor.
At the same time, the reassigned label is drawn from an optimal policy that is learned from reinforcement signals characterized by rules, hence the name \emph{rule-based reinforcement learning}.

\textbf{Physiological Decision Making. }
Recent advances in the healthcare community have shown great promise in automating the laborious process of identifying human physiological states and based on which perform clinical decisions.
Successful examples include whole-night sleep stage reconstruction \cite{sleepCDSS,chenIJCAI,sleep}, epileptic seizures \cite{seizure_ICLR22,kotoge2024splitsee} and arrhythmias \cite{arrhythmias1,arrhythmias2}.
The identification problem is typically cast as a classification framework, under which long-term assessment of patients (e.g. physiological signals) are segmented into raw states of short duration, learned by neural networks for state-relevant features in the hope for capturing long-term dependency of physiological states.
However, beyond the capability of mapping input to physiological states, reliable models must adhere to task-specific rules for clinical decision making
 \cite{CDSSphysiological,AAAIConSequence}.
Physiological phenomena typically exhibit gradual transitions, e.g. the progression from wakefulness to deep sleep \cite{sleeprule}, or the stabilization of heart rates during homeostasis \cite{heartHRV}. 
These sequential transitions are inherently tied to clinical diagnostic criteria \cite{AASM} and must be carefully modeled to enhance the interpretability and physiologically plausibility.

\textbf{Physiological Rules as Learning Constraints. }
Few efforts attempted to embed the physiological rules into the model by e.g. learning rule-conditioned features \cite{Isururule-condition,npjrule-condition,CIKMrule-condition}; incorporating constraints into the loss function \cite{sleeprule2,sleeprule,ChenTNSRE}; employing post-processing modules  to rectify output constraint violations \cite{PostProcess2,PostProcess3,PostProcess1} or integrating additional components to enforce alignment between output and the rules \cite{sleepHMM}. 
Despite a certain degree of effect of these methods, a reliable, effective and general clinical decision support system is still missing in the literature. 
One important reason is that  the rules are domain-specific. Existing methods were typically designed on a case-by-case basis, informed by heavy expert knowledge.
For instance, sleep studies often incorporate rules into representation learning \cite{sleeprule}, whereas works on seizure detection primarily focus on post-processing techniques \cite{PostProcess2,PostProcess1} to smooth the model output and enable holistic analysis. These two methods, each with its pros and cons, are task-specific and cannot be exchanged. Currently, there is no general method that can be utilized across domains, as well as explicit mechanism for rule learning.

\section{Proposed Method}\label{sec:proposed}

Our proposed architecture augments a data-driven model with an additional RL layer.
This layer corrects the physiological impossible predictions from the model by trial-and-error learning and outputs a reassigned label.
This reassignment is evaluated by a simple rule-based reward function given by a known set of impossibility.

\subsection{Problem Formulation}

Assume the classification setting with a dataset $D = \{\bx, y\}$ and a data-driven model $f$.
$\bx$ can be patient profiles or neurophysiological signals.
The model maps input data $\boldsymbol{x}\in\mathbb{R}^{N}$ to one-dimensional class label $ f(\boldsymbol{x}) = \hat{y} \in [K]$, where $K\geq 2$ is an integer  and $[\cdot]$ denotes the integer set $\{1,2,\dots\}$ of all possible labels. 
We use boldface to distinguish between a scalar predicted label $\hat{y}$ from its one-hot vectorized counterpart $\hat{\by}$.
We do not restrict the model class of $f$, though we use a neural network  $f_{\theta}$ in this paper.
The parameters $\theta$ are updated by a loss function that compares the difference between the predicted label $\hat{y}$ to the true label $y$.
In this work, we assume that $f_{\theta}$ has been trained and $\theta$ is frozen, so it does not interact with the RL part.

For each input sample $\bx$, we extract the feature vector from the second-to-last layer of $f_{\theta}$, let us denote it by $\boldsymbol{z} \in \mathbb{R}^{M}$, and typically $M < N$.
The predicted label $\hat{y}_t = f_{\theta}(\boldsymbol{x})$,  feature vector $\bz$ will be used as part of the input to the RL layer, whose goal is to improve prediction accuracy by minimizing the amount of impossible transitions.

\subsection{Markov Decision Process Design}






RL problems are typically formulated as a Markov Decision Process (MDP) defined by the tuple $(\mathcal{S}, \mathcal{A}, P, r, \gamma)$, 
where $\mathcal{S}$ denotes the state space, $\mathcal{A}$ the action space.
$P$ denotes the transition probability and $r$ the reward function.
$\gamma \leq 1$ is the discounting factor that discounts future values. 
We assume the discounting factor to be 1 since in our setting future predictions are of equal importance.

Our RL agent directly works on labels.
Therefore, we define $\mathcal{A}$ as the space of all possible $K$-dimensional one-hot labels.
We distinguish between the one-hot vector $\ba$ and its scalar version $a \in [K]$.
To give sufficient information to the agent to give meaningful reassignment, we design our state space as follows:
at each time step $t$, for input instance $\bx_t$,  we pass it through the classification network to obtain its feature vector $\bz_t$ and the predicted label $\networkMap{\bx_t}$.
Because physiological rules typically depend on consecutive physiological stages, we also remember the previous action or reassigned one-hot label $\ba_{t-1}$.
The three quantities are concatenated to yield a $2K + M$ dimensional state $\bs_t := [\hat{\by}^\top_t, \bz_t^\top, \ba_{t-1}^\top]^\top$.
The agent then samples a new label from its softmax policy $\pi(\ba|\bs)$ to be reassigned to $\bx_t$.

To evaluate the desirability of the state-action pair $(\bs_t, \ba_t)$, we need to define our reward function.
The reward is defined based on the set of impossibility granted by prior knowledge.
For example, we know that a human can never directly transition from wake to rapid eye movement.
Therefore we can simply assign negative reward to any such predicted transition.
To this end, we define $\zeta(a_t)$ as the set of stage labels physiologically one-step reachable from $a_t$.
Depending on whether the RL-agent-reassigned label is correct, the reward is computed by:
\begin{align}
r(\bs_t, \ba_t) = 
    \begin{cases}
        1, & y_t =  a_t \neq \hat{y}_t, \\
        0, & y_t = \hat{y}_t = a_t, \\
        -1, &  y_t \neq \hat{y}_t = a_t,\\
        -2, & y_t \neq \hat{y}_t \neq a_t, \text{ and }  a_t \in \zeta(a_{t-1})\\
        -3, & y_t = \hat{y}_t \neq a_t, \text{ and } a_t \in \zeta(a_{t-1})\\
        -4, & y_t \neq \hat{y}_t \neq a_t  \text{ and } a_t \notin \zeta(a_{t-1}). \\
    \end{cases}
    \label{eq:reward}
\end{align}
In plain words, the reward function gives positive reward only to the RL-corrected predicted labels from wrong to correct, and gives zero reward for \emph{maintain and correct} labels.
For the negative part, the rule is slightly finer-grained:
$-1$ reward is incurred when the agent does not change a wrong prediction.
$-2$ and $-3$ reward are respectively incurred when the predicted label is correct/wrong, and the selected action is wrong but physiologically possible.
The worst case $-4$, is given as penalty when the action is both wrong and physiologically impossible.

It can be seen the reward function is general and the only part that needs prior knowledge is the set of impossibility $\zeta(a_{t-1})$.
This set can be much smaller in general than the set containing all possible transitions, and requires much lighter expert knowledge in contrast to existing work \citep{Saboo2021-RLDiseaseProgresionModelAlzheimer,Fatemi2021-RLidentifyMedicalDeadends,Hu2023-REMEDI-ML4H-RLadaptiveMEtabolismmodeling}.
Another point worth noting is that,
the true label information is not given to the agent directly.
Rather, it is implicitly presented to the agent to infer the best action.
When the RL layer is trained sufficiently on the training data, it is hoped that the layer can generalize to the test data where the true label information is no longer available.


\subsection{Rule-based Reinforcement Learning Layer}

It is clear that the problem tackled by the RL layer depends on the quality of the preceding data-driven model.
In the extreme case where the model is perfect with 100\% accuracy, there is nothing can be learned by the RL layer and the highest possible cumulative return is simply $0$.
On the other hand, if the model is poor, it gives a large room for potentially higher cumulative reward. 
But in this case, since the predicted label is highly likely to be wrong, its feature vector $\bz$ may not be very useful, and the RL agent would rely on randomly selecting other labels until bumping into the correct one, which requires longer training horizon and/or stronger exploration mechanism.
To compensate for this, we adopt the classic $\epsilon$-greedy method to facilitate exploration.

Let us define the cumulative return $ \sum_{t=0}^{T} r(\bs_t, \ba_t)$ summing over the prediction horizon $T$.
Our goal is to learn an optimal policy that maximizes the expected return $G_t := \expectation{}{\sum_{t=0}^{T} r(\bs_t, \ba_t)}$.
To train the layer, we can let $T$ denote the length of the trajectory input to the classifier network $\networkMap{\cdot}$.
However, when the trajectory is long, it may be more preferable to manually segment the input e.g. by individuals so that $T$ becomes smaller.

Among the plethora of RL algorithms, the classic policy search algorithm REINFORCE \citep{Williams-REINFORCE} stands out for our setting due to its simplicity and interpretability.
REINFORCE is an on-policy algorithm that optimizes the policy towards maximizing cumulative reward by sampling from the current policy \citep{Sutton1999}.
This fits well with the healthcare setting where it is typical to iterate each individual sample to update the policy \citep{Hartvigsen2019-EARLIEST}.
Extending the framework to off-policy or even offline algorithms is an interesting future direction.

Let us denote the policy network parametrized by $\phi$. We optimize the network by minimizing the following loss:
\begin{align*}
    J(\phi) \!:=\! -\expectation{\!}{ \sum_{t=0}^{T} \log \pi_{\phi} (\ba_t|\bs_t) \!\adabracket{ \sum_{t'=t}^{T} r(\bs_{t'}, \ba_{t'}) - b_{\psi}(\bs_{t'}) \!\!}\!},
\end{align*}
where $b_{\psi}$ is a state-dependent baseline parametrized by $\psi$ for better numerical stability.
The baseline network is simply updated by regression to the summation of rewards 
\begin{align*}
    J(\psi) := \expectation{}{\adabracket{b_{\psi}(\bs_{t'}) - \sum_{t'=t}^T r(\bs_{t'}, \ba_{t'})}^2 }.
\end{align*}
In physiological processes, it is typical for a status to remain unchanged for a period of time.
To take this fact into account, we impose a penalty when the current action is different than the last action.
In summary, RRLL optimizes the following objective:
\begin{align}
    \mathcal{L} = J(\phi) + J(\psi) + \alpha \sum_{t=0}^{T}  \mathbbm{1}\{\ba_t \neq \ba_{t-1}\} 
 \log\pi(\ba_t | \bs_t),
\end{align}
where $\alpha > 0$ is a weighting coefficient for the penalty.
Minimizing the likelihood is equivalent to minimizing the probability of an action that leads to a reassigned label.
After the training of RRLL, the entire model can be used as a  whole to perform end-to-end inference with RRLL acting simply as the last layer for predicting labels.
Implementation details are provided in Appendix \ref{apdx:implementation}.

\subsection{Workflow of RRLL Training}

Figure \ref{fig:overview} illustrates the workflow of training RRLL. 
It is worth pointing out that the base predictor does not interact with RRLL in the sense that once its weights are frozen once it has been trained.
Therefore, RRLL relies only on the predicted labels and  feature vectors.
This renders RRLL lightweight and general since only labels and features need to be stored but not the potentially complex model that could comprise billions of parameters.
RRLL draws a new label as its action to the input instance and the input state and action will be evaluated by the set $\zeta$.
After training, the entire model (base predictor plus RRLL) can be used end-to-end on any prediction task same as the base predictor.

\textbf{Remark. }
On the technical level, our method bears some similarity to \citep{Peng2018-FastDiseaseDiagnosisbyRL-Refuel,Hartvigsen2019-EARLIEST}.
Specifically, \citet{Peng2018-FastDiseaseDiagnosisbyRL-Refuel} used binary vectors representing symptoms as input and outputs a binary vector indicating potential diseases.
By contrast, we gather one-hot predicted labels, previous action and features to give sufficient information to the agent to output sensible relabeling, acting as a teacher to the classifier.
\citet{Hartvigsen2019-EARLIEST} also adopted REINFORCE with similar losses for learning an early-stopping seizure detection policy.
Their setting is simpler as the agent faces a sequence of binary classification problems of stop or continue, and each trajectory yields either 0 or 1 reward.
RRLL by contrast, needs to simultaneously discern physiologically impossible actions from a sequence of rewards defining different situations at each step and reassign a correct label.


\section{Experiments}\label{sec:experiment}

We verify the utility of RRLL by important healthcare classification problems that often see physiological impossible predictions.
We take the state-of-the-art models as the base predictor and investigate how RRLL can improve prediction by reducing physiologically impossible predictions.


\subsection{Proof-of-Concept: Sleep Staging}

Human sleep goes through several sleep stages from light sleep to deep sleep.
Automating the laborious manual sleep staging process has been a popular topic in the AI4Science literature.
However, using RL for better stage prediction has not been studied before to the best of our knowledge.
We use the largest public sleep database -- the Sleep Heart Health Study\footnote{\url{https://sleepdata.org/datasets/shhs}}that has 42,560 hours of EEG recordings from 5,793 subjects.
Sleep stages are scored into 5 classes: Wake, N1, N2, N3 and REM.
We use 85\% of patients for training and 15\% for testing.

\begin{table}[t]
\centering
\caption{Comparison of performance on the SHHS sleep dataset.
RRLL on top of the base predictor attains the highest performance among all baselines.
}
\resizebox{\linewidth}{!}{%
\begin{tabular}{l|ccccccc} 
\toprule
Models & Metric & Wake & N1 & N2 & N3 & REM & $k$ \\ 
\midrule
\multirow{2}{*}{AttnSleep} & Pre & 0.90 & 0.30 & 0.87 & 0.87 & 0.80 &  \\
& Re & 0.83 & 0.36 & 0.86 & 0.87 & 0.83 &0.81 \\
\cite{emadeldeen} & F1 & 0.86 & 0.33 & 0.87 & 0.87 & 0.82 & \\ 

\midrule
\multirow{2}{*}{SleepFormer } & Pre & 0.93 & 0.51 & 0.89 & 0.83 & 0.90 &  \\
& Re & 0.92 & 0.42 & 0.87 & 0.86 & 0.86 &0.84 \\
\cite{sleeprule}& F1 & 0.92 & 0.46 & 0.88 & 0.85 & 0.88 & \\ 

\midrule
\multirow{2}{*}{FC-Attention } & Pre & 0.93 & 0.42 & 0.87 & 0.89 & 0.80 &  \\
& Re & 0.93 & 0.33 & 0.89 & 0.84 & 0.79 &0.80 \\
\cite{ChenTNSRE}& F1 & 0.93 & 0.38 & 0.88 & 0.87 & 0.80 & \\ 

\midrule
\multirow{2}{*}{FC-Attention +} & Pre & 0.94 & 0.50 & 0.89 & 0.89 & 0.88 &  \\
& Re & 0.93 & 0.44 & 0.90 & 0.85 & 0.85 &0.85 \\
SleepFormer& F1 & 0.93 & 0.47 & 0.89 & 0.87 & 0.86 & \\ 

\midrule
\multirow{2}{*}{Ours } & \cellcolor{platinum}Pre & \cellcolor{platinum}0.97 & \cellcolor{platinum}0.70 & \cellcolor{platinum}0.92 & \cellcolor{platinum}0.90 & \cellcolor{platinum}0.92 & \cellcolor{platinum} \\
& \cellcolor{platinum}Re & \cellcolor{platinum}0.97 & \cellcolor{platinum}0.58 & \cellcolor{platinum}0.94 & \cellcolor{platinum}0.90 & \cellcolor{platinum}0.94 & \cellcolor{platinum}0.88 \\
(FC-Attention + \textbf{RRLL})& \cellcolor{platinum}F1 & \cellcolor{platinum}0.97 & \cellcolor{platinum}0.64 & \cellcolor{platinum}0.93 & \cellcolor{platinum}0.90 & \cellcolor{platinum}0.93 & \cellcolor{platinum}\\ 
\bottomrule 
\end{tabular}
}

\label{tab:sleep_comparison}
\end{table}

\begin{figure*}[th]
    \centering
        \includegraphics[width=0.9\textwidth]{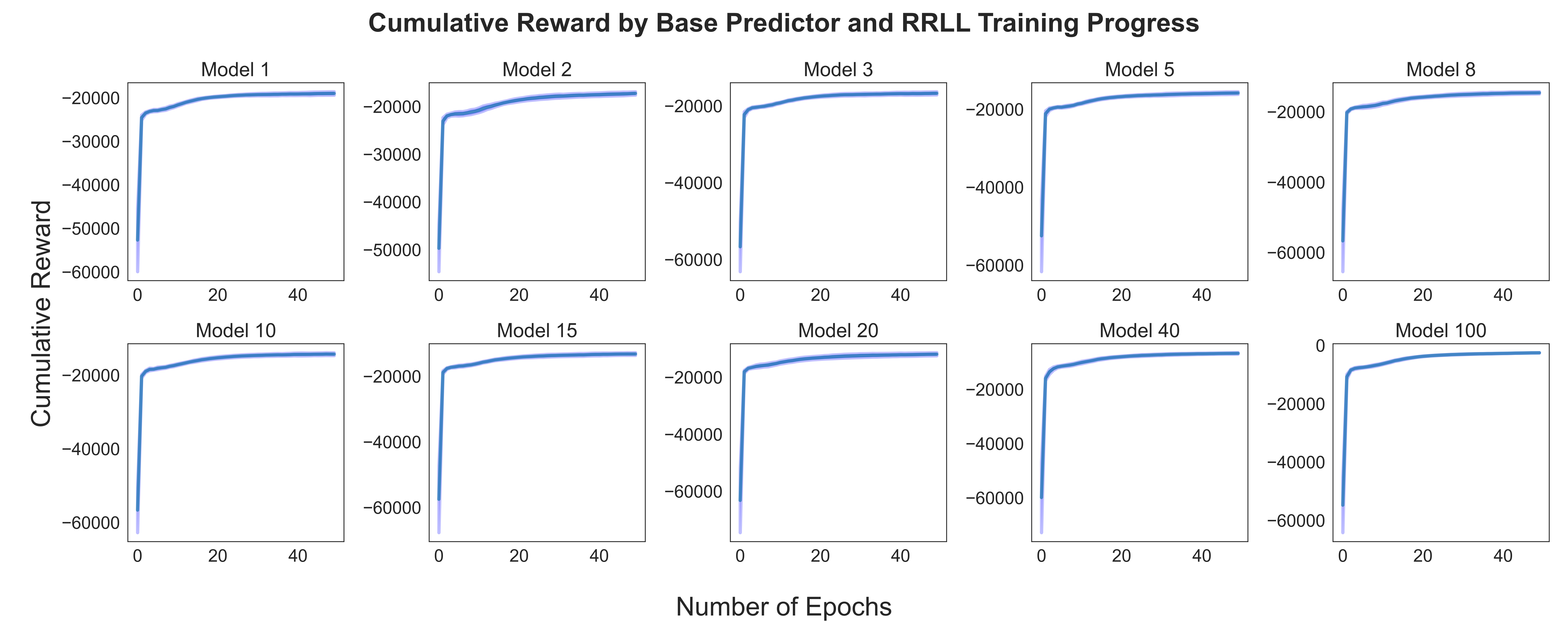}
        \vspace{-10pt}
    \caption{Cumulative reward on all base predictor models, averaged over 10 seeds.
    Model index is the the base predictor model trained after model index epochs.
    Ribbons show the 95\% confidence interval.
    }
    \label{fig:reward}
\end{figure*}

\begin{figure}[th]
    \centering
    \includegraphics[width=0.92\columnwidth]{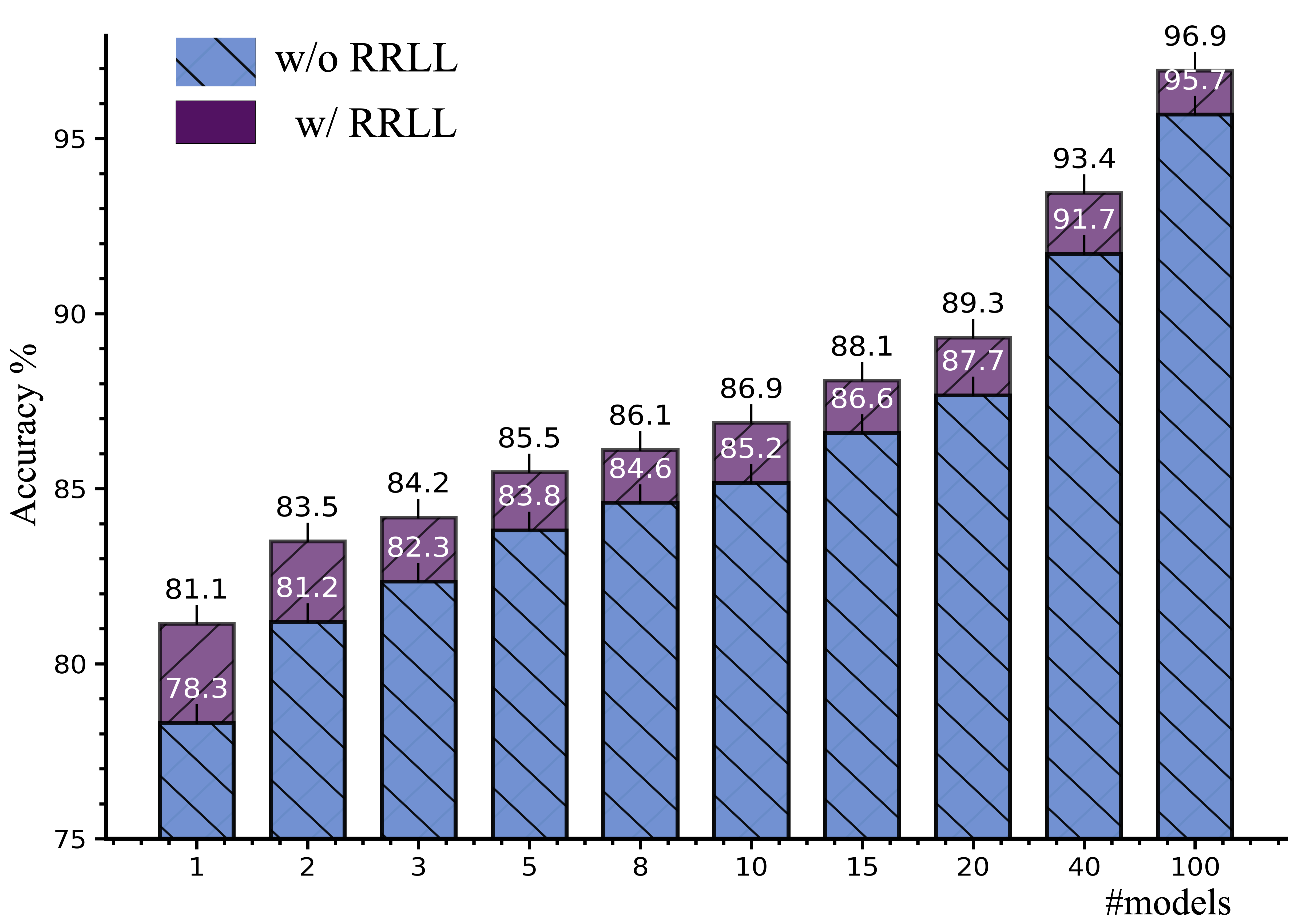}
    \caption{
    RRLL is capable of improving the accuracy of all base models, showing the general applicability of RRLL.
    }
    \label{fig:bar}
    \vspace{-5pt}
\end{figure}

\begin{figure}[th]
    \centering
\includegraphics[width=0.96\columnwidth]{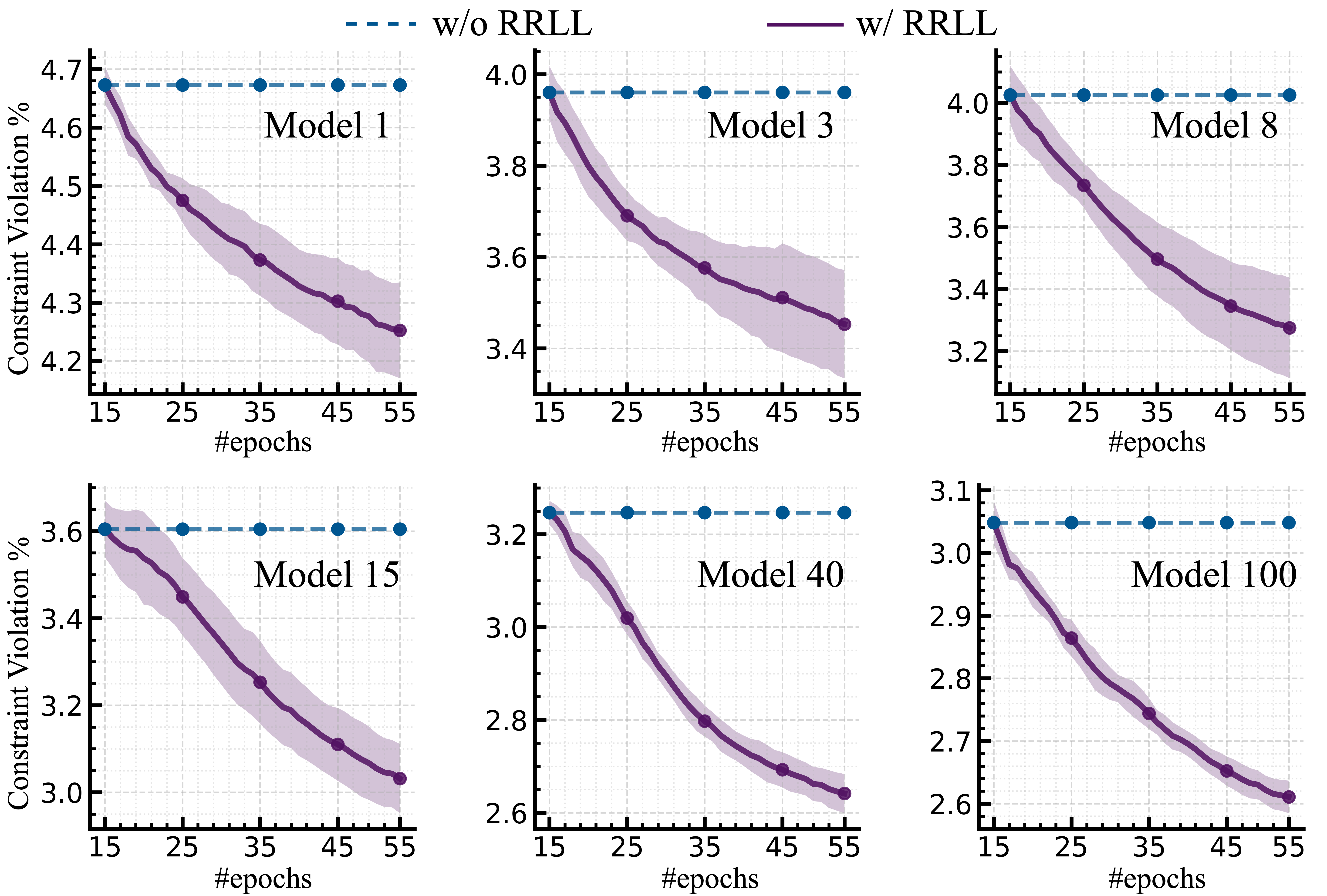}
    \caption{
    The percentage of physiologically impossible predictions along RRLL learning. Ribbons show the 95\% confidence. The blue dashed lines show the base predictor.
    Though the base predictor percentage decreases from Model 1 to 100, RRLL can still notably reduce it on Model 100, indicating that the base predictor does not improve accuracy by explicitly minimizing physiologically impossible predictions.
    }
    \vspace{-5pt}
    \label{fig:cv}
\end{figure}

We choose the fragment-wise FC-Attention model \citep{ChenTNSRE} as the base predictor.
To evaluate RRLL along during the base predictor training, 
we store the features and predicted labels ($\bz, \hat{y}$) after a certain number of training epochs (1, 2, \dots 100) and label these models by the number of epochs they have been trained, e.g. model 100 indicates the model trained after 100 epochs.
RRLL is then applied to each of the models.
This allowed us to evaluate the effectiveness of RRLL from both the predictor training and the RL learning axes.
In terms of evaluation metrics, Precision (Pre), Recall (Re) and F1-score are adopted.
As for the baseline algorithms to be compared with RRLL + predictor, we opt for the following four baselines: AttensionSleep \citep{emadeldeen}, SleepFormer \citep{sleeprule}, FC-Attention \citep{ChenTNSRE}, and FC-Attention+SleepFormer.

\textbf{Sleep Rules. }
The five distinctive sleep stages from wake to light stages (N1 and N2) and deep sleep (N3), culminating in rapid eye movement (REM) \cite{AASM} represent gradual, transitioning processes, standardized into clinical rules to guide experts in labeling whole-night physiological recordings (e.g., EEGs) \cite{EEGscience}.
Though it may be rare but still possible that a human goes from REM to wake; going from wake to REM is impossible.
Therefore, for sleep we can define the set of impossibility $\zeta$:
\begin{align}
    \zeta_\texttt{sleep} :=
    \begin{cases}
        \text{Wake}  &\not\rightarrow \quad \text{N3, REM}\\
        \text{N1} & \not\rightarrow \quad \text{N3, REM}\\
        \text{N2} & \not\rightarrow \quad \text{Wake}\\
        \text{N3} & \not\rightarrow \quad \text{N1}\\
        \text{REM} & \not\rightarrow \quad \text{N1, N3}.
    \end{cases}
    \label{eq:rule_sleep}
\end{align}

This set of rules is used in computing the reward Eq. (\ref{eq:reward}).
Whenever the agent tries to take an action $a_t$, it will be judged by the rules to see if it is a possible transition by $ \zeta_\texttt{sleep}(a_{t-1})$.
An impossible transition will be counted as a constraint violation.
We visualize how RRLL improves accuracy by reducing constraint violation in the results.



\begin{figure*}[th]
    \centering
\includegraphics[width=0.97\textwidth]{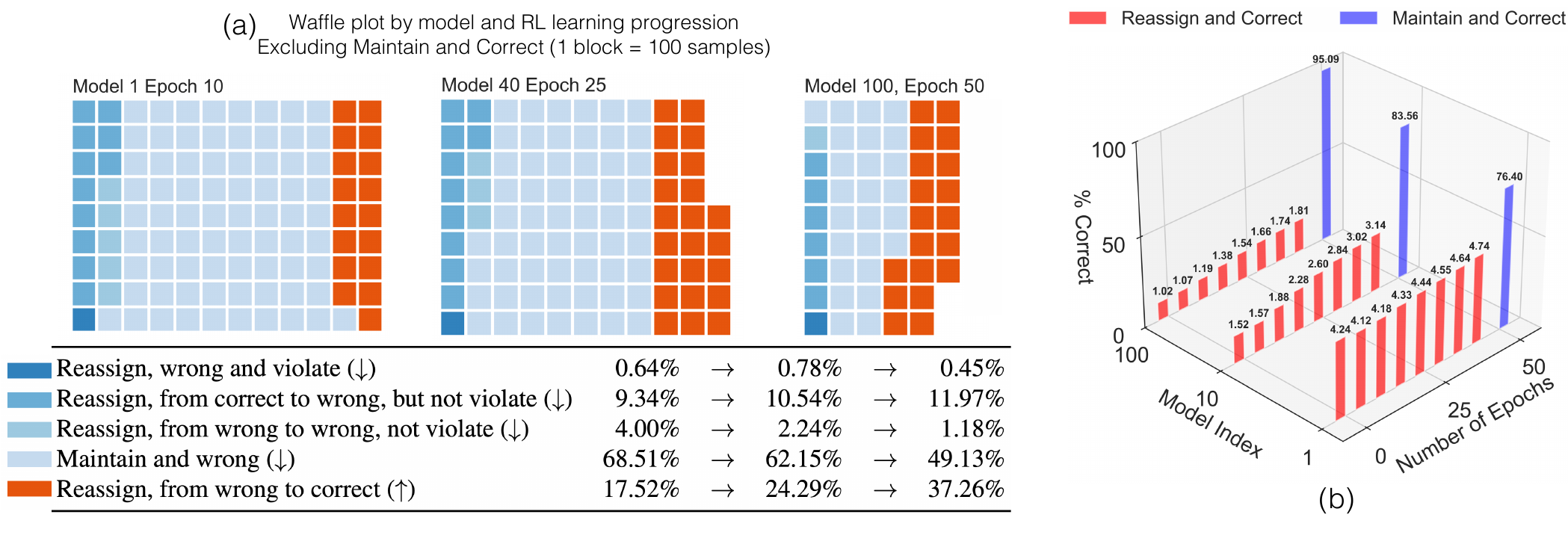}
    \vspace{-10pt}
    \caption{
    (Left) Waffle plot of class instances defined by the reward function Eq. (\ref{eq:reward}) along the base predictor model and RRLL training. The class \emph{Maintain and Correct} is excluded since it is significantly larger than others. 
    As a result of learning, the totality of the five classes shrinks, and the percentage of successful reassignment increases, with that of the wrong attempts decreases. 
    (Right) RRLL correct assignments consist of two cases: Maintain and Reassign.
    Both classes steadily become larger along with model and RRLL learning. At the end they account for $\approx 97\%$ accuracy.
    }
    \vspace{-5pt}
    \label{fig:category}
\end{figure*}

\begin{figure*}
    \centering
        \includegraphics[width=0.95\textwidth]{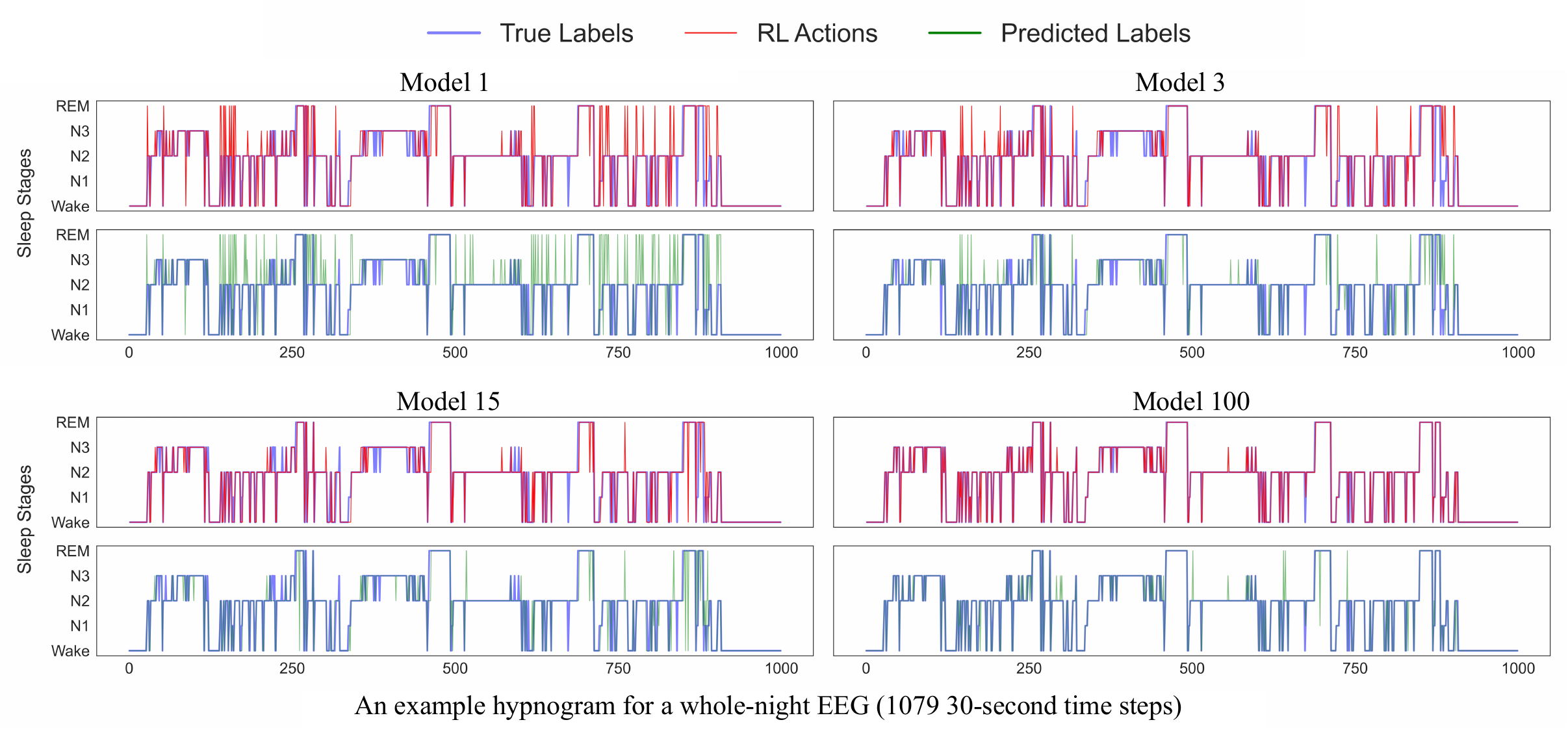}
    \caption{
    RRLL can effectively reduce physiologically impossible transitions along the way of base predictor training, even on Model 100 where the base predictor's predictions are already accurate with no much room for further improvement.
    }
    \label{fig:sleepvis}
\end{figure*}

\begin{figure}
    \centering
\includegraphics[width=0.99\columnwidth]{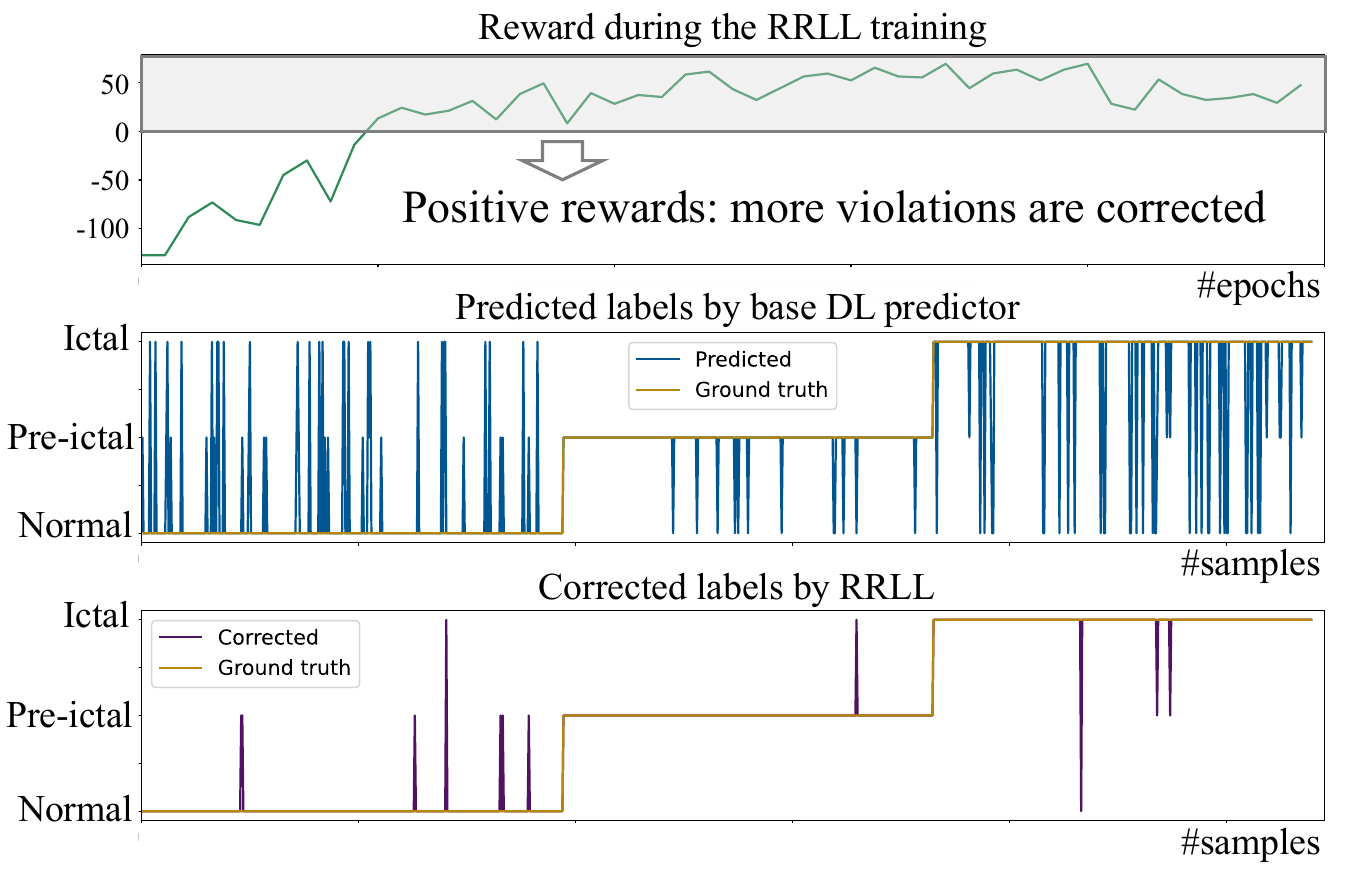}
    \vspace{-10pt}
    \caption{
    RRLL on the challenging seizure prediction task.
    The base predictor outputs highly oscillating predictions. RRLL by contrast is capable of removing most of the impossible transitions between Ictal and Normal.
    }
    \vspace{-5pt}
    \label{fig:seizure}
\end{figure}


\textbf{Results. }
Table \ref{tab:sleep_comparison}  summarizes the performance of RRLL on sleep.
It can be seen that RRLL outperforms all baselines in both tasks by a noticeable margin.
From the model learning perspective, RRLL contributes a new, state-of-the-art model that can be readily used for any prediction task.


To analyze why RRLL can achieve the superior performance, we conduct extensive visualization.
Figure \ref{fig:reward} shows the cumulative reward learning curves of RRLL on top of all base predictor models from 1 to 100.
It is visible that RRLL is capable of quickly and robustly converging on all base models. 
Since the reward function Eq. (\ref{eq:reward}) encodes the desirability of each state, a rise in cumulative rewards indicates an increase of correct assignments and a decrease of physiological impossible predictions.
The increase of accuracy can be seen from figure \ref{fig:bar} where the average improvement attained by RRLL is shown to be consistently better than the original model from the very beginning of learning, to the point where the model starts to overfit (brown bars of model 40 and 100).
This shows RRLL is robust even in the extreme case even when the base predictor model is highly accurate with no much room for further improvement.

Figure \ref{fig:cv} visualizes how RRLL reduces the percentage of constraint violation per Eq. (\ref{eq:rule_sleep}) in the entire dataset along with the model learning and RL progress.
By observing model 100, another observation can be drawn: the predictor model improves accuracy not by minimizing physiological impossible predictions, which still account for a non-negligible portion in model 100 and therefore can still be effectively reduced by RRLL.

Figure \ref{fig:category} LHS shows a Waffle plot \citep{pywaffle} of samples associated with different rewards,  excluding the action \emph{maintain and correct}, which accounts for the majority of samples.
It is clear that along with learning, the totality of wrong assignments shrinks, and the proportion of RRLL correct reassignments increases.
Figure \ref{fig:category} RHS shows the totality of RRLL correct assignments: the proportion of successful actions, no matter maintain or reassign, increases along the RL epochs (the action maintain is only show for the last epoch for uncluttered visualization).
At the end of learning, they in total accounts for $\approx 97\%$ accuracy.

To gain more intuition of the effectiveness of RRLL, we showcase with a whole-night sleep hypnogram of a selected patient. 
It can be seen that even on Model 1, RRLL can notably reduce impossible transitions displayed as the green peaks such as the N2-to-REM misclassifications.
Even as the base predictor becomes more accurate, RRLL is still capable of removing constraint violations, as can be seen from Model 100, between step 250 and step 750.

\subsection{Extra Validation: Seizure Onset Detection}

Another important task that suffers from discontinuous and impossible predictions is seizure onset detection.
Around 10\% of people will experience a seizure during their lives \citep{Devinsky2018-EpilepsyReviewPrimer}.
Accurate seizure onset prediction from electrophysiological changes is immensely helpful in providing timely intervention.
However, accurate prediction is still challenging and existing results remain limited.

We use the CHB-MIT database\footnote{\url{https://physionet.org/content/chbmit/1.0.0/}} that has 844 hours of 22-channel scalp EEG data from 22 patients, including 163 recorded seizure episodes.
In this example, three stages are defined as Normal, Preictal (before the actual seizure onset) and Ictal (seizure onset).
We opt for the one of the state-of-the-art models \cite{seizure_ICLR22} as our base predictor.
For the baselines, we opt for the two-stage post-processing methods that reassign labels to sequences of EEG fragments based on the outputs of prediction models \citep{PostProcess2,PostProcess1}.
 The first baseline called Post  employs label voting within a sliding window.
 The second baseline ScoreNet utilizes LSTM to further score the labels and their corresponding latent vectors.
We use the following well-established metrics for seizure onset detection to evaluate performance: normalized mutual information (NMI), adjusted Rand index (ARI), and accuracy (ACC).

\begin{table}[t]
    \centering
    \caption{Performance on the CHB-MIT seizure dataset.
    }
    \vspace{3pt}
    \resizebox{0.99\linewidth}{!}{%
    \begin{tabular}{l|lll}
    \toprule
    Method & NMI & ARI & ACC \\ \midrule
    DCRNN   & 0.865 $\pm$ 0.07 & 0.867 $\pm$ 0.08 & 0.939 $\pm$ 0.08 \\
    DCRNN + Post     & 0.897 $\pm$ 0.04 & 0.875 $\pm$ 0.03 & 0.882 $\pm$ 0.03 \\
    DCRNN + ScoreNet     & 0.842 $\pm$ 0.03& 0.744 $\pm$ 0.03& 0.868 $\pm$ 0.02\\

    DCRNN + \textbf{RRLL}             & \textbf{0.979} $\pm$ 0.01 & \textbf{0.964} $\pm$ 0.01 & \textbf{0.981} $\pm$ 0.01\\
    \bottomrule
    \end{tabular}
    }
    \label{tab:seizure_chbmit}
\end{table}

\textbf{Seizure Rules. }
The rules for seizure progression are straightforward: patients transition from a normal state to a pre-ictal state and finally to an ictal (seizure) state.
In the dataset we do not take ictal to normal into account:
\begin{align*}
    \zeta_\texttt{seizure} :=
    \begin{cases}
        \text{Normal}  &\not\rightarrow \quad \text{Ictal}\\
        \text{Preictal} & \not\rightarrow \quad \text{Normal}\\
        \text{Ictal} & \not\rightarrow \quad \text{Normal, Preictal}
    \end{cases}
\end{align*}
It is worth noting that because the task is hard and the dataset is imbalanced, the predictor outputs wrong predictions that make RRLL learning slow.
We find that a simplified reward function that removes the condition on the predicted label helps the agent better learn:
\begin{align*}
    r(\bs_t, \ba_t) =    
    \begin{cases}
        1, & a_t = y_t \neq \hat{y}_t \\
        0, & a_t = y_t  = \hat{y}_t \\
        -1, & a_t \neq y_t \text{ and } a_t \in \zeta(a_{t-1}) \\
        -2, & a_t \neq y_t \text{ and } a_t \notin \zeta(a_{t-1})
    \end{cases}
\end{align*}
In plain words, positive rewards is given to successful corrections, and negative rewards depend only whether or not an action is correct and violates the rules.

\textbf{Results. }
Table \ref{tab:seizure_chbmit} summarizes the performance of the baselines. 
It is clear that RRLL with the base predictor again attains a significant edge.
Figure \ref{fig:seizure} examines the improvement more carefully by visualizing a selected patient going through going through the Normal-Preictal-Ictal process.
Since in the considered dataset it is not possible for a patient to jump between Normal and Ictal, a transition between the two states is considered a constraint violation or impossible prediction.
The task is known to be challenging, as can be see from the second window that the base predictor outputs highly oscillating predictions alternating between the three possible states. 
By contrast, from the third window it is evident RRLL is capable of removing most of the physiologically impossible predictions that jump from ictal to normal and vice versa. 
This can also be seen from the top window showing cumulative rewards.
Since positive reward is only given to successful corrections, positive cumulative reward indicates the agent has learned to maintain correct predictions and minimize the impossible ones.

\section{Conclusion}

This paper proposed Rule-based Reinforcement Learning Layer (RRLL) that augments any base predictor with the capability of correcting physiologically impossible predictions.
RRLL is a layer as it allows the entire model to predict still in an end-to-end manner.
At the same time, it is learned in an RL fashion signaled by rules: it takes in as states predicted labels and outputs reassigned labels, evaluated by a reward function based on a small set of impossible transitions.
RRLL complements the existing literature on RL for healthcare, and provides a workable Markov Decision Process design that has been shown to perform favorably with a classic policy search algorithm.
Extensive experiments showed that RRLL was capable of improving prediction accuracy by effectively reducing impossible predictions, which was not possible by training the base predictor alone.

One interesting future possibility lies in trying more advanced RL algorithms such as the off-policy or even offline methods.
Another promising direction is to design a more general Markov Decision Process such that the domain-specific set of impossibility can be removed.




\clearpage

\section*{Impact Statement}


This paper focuses on improving prediction of machine learning models on physiological problems. 
The data used in the paper comes from public sources, and our model falls into the regular deep learning models. 
While our model has the potential to facilitate automated diagnosis/prediction, we feel the societal consequences are not specifically needed to be highlighted here.


\bibliography{example_paper}
\bibliographystyle{icml2025}

\newpage
\appendix
\onecolumn
\section*{Appendix}

\section{Experimental Setting}\label{apdx:experiment_setting}

\subsection{Sleep Staging}

\textbf{Baselines. }
We selected four baselines: AttnSleep \cite{emadeldeen}, SleepFormer \cite{sleeprule}, FC-Attention \cite{ChenTNSRE}, and FC-Attention+SleepFormer.
Clinically, sleep staging is performed on 30-second EEG fragments.
AttnSleep and FC-Attention are fragment-wise models that label each fragment individually. 
They focus on capturing stage-specific temporal or frequency characteristics (e.g., 12–16 Hz sigma waveforms in N2 \cite{wake}) according to AASM rules, enabling more physiologically meaningful staging outcomes.
SleepFormer follows a sequence-to-sequence paradigm \cite{seq-to-seq}, taking multiple EEG fragments as input to Transformer models. 
It explicitly learns sequential transition information and simultaneously labels a sequence of EEG fragments.
\cite{ChenTNSRE} also evaluate FC-Attention+SleepFormer in their work, which combines the intra-fragment stage-specific characteristics of FC-Attention with inter-fragment transition learning using sequence-to-sequence approach.

\textbf{Setup and Metrics. }
We selected the fragment-wise FC-Attention model as the predictor for our experimental settings.
FC-Attention, a Transformer-only general model, effectively captures intra-fragment characteristics but does not account for inter-fragment or stage transition information in its learning process.
In our approach, we first trained the predictor to sequentially output labels for each 30-second EEG fragment in patient recordings.
We then applied our RRLL framework to reassign these predicted labels based on rule learning.
During predictor training, we recorded ($\bz, \hat{y}$) across different training epochs (e.g., 1, 2, and 20, with early stopping applied at the best performance) and applied RRLL to these models.
This allowed us to evaluate the effectiveness of RRLL from an untrained state to a well-converged model.
We use the well-established metric: precision (Pre), recall (Re) and F1-score to evaluate the effectiveness of baselines and RRLL on SHHS.

\subsection{Seizure Onset Detection}

\textbf{Dataset. }
We use the CHB-MIT database that has 844 hours of 22-channel scalp EEG data from 22 patients, including 163 recorded seizure episodes.
Following the work of \cite{kotoge2024splitsee}, we defined the five-minute period before a seizure as the preictal phase. 
We divided each dataset into 70\%/25\%/5\% for training, testing, and validation. 
Patient recordings were segmented into 1-second fragments, and the IDs of all fragments were stored to enable recall in long-term recordings for verifying RRLL.

\textbf{Baselines. }
We compared our proposed method with two post-processing seizure prediction baselines, both of which are two-stage methods that re-assign consistent state labels to sequences of EEG fragments based on the outputs of prediction models.
\cite{PostProcess2} used a sliding 5-second window to re-assign labels based on a patient-specific voting threshold.
\cite{PostProcess1} applied a weighting phase to score the probabilities of seizure onsets.

\textbf{Setup and Metrics. }
Since the two baselines primarily focus on second stage post-processing,
for the predictor in both baseline methods and our proposed setting, we selected the state-of-the-art DCRNN model \cite{seizure_ICLR22}.
DCRNN uses 12-second EEG fragments as input, effectively capturing brain state changes over time.
We then applied the two post-processing baselines and our RRLL framework to correct the model-predicted labels.
We evaluated performance of state partitioning using two clustering metrics: normalized mutual information (NMI) and adjusted Rand index (ARI), and accuracy (ACC) for effectiveness of seizure onset detection.

\textbf{Rules. }
Approximately 40\% of epileptic patients have drug-resistant epilepsy with recurrent seizures that cannot be controlled by available medications \cite{PostProcess1}.
Clinically, many patients benefit from accurate state partitioning of EEG recordings (e.g., normal, pre-ictal, or ictal states) and seizure onset detection, as these help localize and surgically remove the onset zone in the brain, which exhibits the earliest electrophysiological changes during a seizure event.
However, current fragment-wise classification models focus on accurately classifying individual fragments without explicitly modeling long-term dependencies or state transitions. This often results in abrupt misclassifications within otherwise state-consistent sequences.
In general, the rules for seizure progression are relatively straightforward: patients transition from a normal state to a pre-ictal state and finally to an ictal (seizure) state, as follows:

\section{Implementation Details}\label{apdx:implementation}


\begin{table}[t] 
\begin{small}
\begin{center}
\caption{Parameters for RRLL. }
\begin{tabular}{lcccccc}
\toprule
\textbf{Parameter} & \textbf{Value} \\
\midrule
Learning rate & \makecell{Swept in $\{3\times 10^{-5}, 3\times 10^{-4}, 3\times 10^{-3}\}$   }\\[1ex] \hline
Smooth coefficient $\alpha$ & \makecell{Swept in $\{10, 1.0, 0.1\}$ } \\ \hline
Temperature of the softmax policy & Swept in $\{10, 1.0, 0.1\}$\\ \hline
Epsilon greedy $\epsilon$ & Swept in $\{0.5, 0.1, 0.01\}$ \\ \hline
Training Epochs & 50 \\ \hline
Hidden size of policy network  & $32 + 2 \times K$  \\ \hline
Hidden layers of policy network & 2 \\ \hline
Hidden size of baseline network & $32$ \\ \hline
Hidden layers of baseline network & 2 \\ \hline
Optimizer & Adam \\ \hline
Scheduler & Exponential learning rate scheduler \\ \hline
Exponential lr discount factor & $0.99$ \\ \hline
Adam.$\beta_1$ & 0.9 \\ \hline
Adam.$\beta_2$ & 0.99 \\ \hline
Number of seeds for sweeping & 10 \\ 

\bottomrule
\end{tabular}
\end{center}
\end{small}
\label{tab:hyper}
\end{table}

\textbf{RRLL.}
The hyperparameters for RRLL is summarized in Table \ref{tab:hyper}.
Specifically, the policy is a discrete softmax distribution that transforms input state $\bs$ to $K$-dimensional logits $\bq$, remember that $K$ is the number of classes:
    $\pi(\ba_i|\bs) = \frac{\exp \adabracket{\frac{\bq_i}{\eta}}}{\sum_{i=1}^{K} \exp\adabracket{\frac{\bq_i}{\eta}}}$,
where $\bq_i$ is the $i$-th logit and $\eta > 0$ is the temperature coefficient. 
Since we use the $\epsilon$-greedy method, at each step, we generate a random number that if larger than $\epsilon$, we sample from the policy $\pi$, otherwise uniformly random from $[K]$.
We use the Adam optimizer with exponential learning rate scheduler with discount factor $0.99$.
The learning rate, smooth coefficient $\alpha$, temperature coefficient $\eta$ and epsilon-greedy $\epsilon$ are swept. 
The policy network has latent dimension of $32 + 2\times K$, where $2K$ is the dimension of one-hot predicted label and last action.

\begin{table}[t]
\caption{Parameters for sleep staging base model.}
\centering
\begin{tabular}{l|c}
\toprule
\textbf{Parameter}                        & \textbf{Value}   \\ \midrule
\#Stacked encoder                         & \{6, {8}, 12\} \\
\#Heads ($h$)                         & \{2, 4, {8}, 12\}      \\
Dimension of linear projection of $D$ & \{16, {32}, 64\}   \\
Normalization-like scale ($\sqrt{d}$)               & \{2, {4}\}        \\
Dimension of MLP output                          & \{64, {128}, 256\} \\
Dropout rate                              & \{0.2, {0.5}, 0.8\} \\
\#Training epoch                              & 100 \\
Batch size                              & 32\\
\#Parameters                              & $1.3 \times 10^{5}$\\
\bottomrule
\end{tabular}
\label{tb3}
\end{table}

\textbf{Sleep Staging Implementation. }
To address data imbalance, we implement a pre-training-to-fine-tuning strategy.
Additionally, we conduct a two-phase training process, where the model with the best parameter settings is passed to RRLL.
In the first phase, we determine the optimal parameter settings through grid search.
Then, we re-train the model, record checkpoints at different parameter settings, and generate tentative prediction labels for RRLL to evaluate its effectiveness, as illustrated in Figure \ref{fig:category}.
To mitigate performance overestimation, we employ subject-wise 7-fold cross-validation, splitting the dataset into seven subject-wise subsets.
In each trial, six subsets are used for training, while the remaining subset (approximately 800 subjects) is used for validation.
All experiments are conducted on a server equipped with NVIDIA RTX A6000 GPUs.


\paragraph{Seizure Detection Implementation.}

We used binary cross-entropy as the loss function to train all models.
The parameter settings strictly followed those in \citep{seizure_ICLR22}.
During training, we applied data augmentation by randomly scaling the amplitude of raw EEG signals by a factor between 0.8 and 1.2.
The models were trained for 100 epochs with an initial learning rate of 1e-4.
To enhance efficiency and sparsity, we set $\tau = 3$, retaining only the top-3 nearest neighbor edges for each node.
The dropout probability was set to 0 (i.e., no dropout).
The base model consists of two GRUs with two stacked layers and a two-layer GCN with 64 hidden units, resulting in 114,794 trainable parameters.
We compared selected baselines to RRLL. 
For a fair comparison, with \cite{PostProcess2} and \cite{PostProcess1}, we maintained our first-stage base predictor and compared their post-processing methods.

\end{document}